\def\year{2022}\relax
\documentclass[letterpaper]{article} 
\usepackage{lineno}
\usepackage{makecell}
\usepackage{aaai22}  
\usepackage{amsmath}
\usepackage{blindtext}
\usepackage{graphicx}
\usepackage{amssymb}
\usepackage{graphicx}
\usepackage{bbm}
\usepackage{mathtools}
\usepackage{mathrsfs}
\usepackage{amsthm,amsmath,amssymb}
\usepackage{mathrsfs}

\usepackage{multirow}
\usepackage{algorithm}
\usepackage{algorithmic}
\usepackage{bbding}
\usepackage[table,xcdraw]{xcolor}
\usepackage{epsfig}

\usepackage{amsmath}
\usepackage{subfigure}
\usepackage{times}
\usepackage{epsfig}
\usepackage{graphicx}
\usepackage{amsmath}
\usepackage{amssymb}
\usepackage{bm}
\usepackage{amsfonts,amssymb}
\usepackage{booktabs}
\usepackage{float}

\usepackage{subfigure}
\usepackage{makecell}
\usepackage{multirow}

\usepackage{times}  
\usepackage{helvet}  
\usepackage{courier}  
\usepackage[hyphens]{url}  
\usepackage{graphicx} 
\usepackage{natbib}  
\usepackage{caption} 
\DeclareCaptionStyle{ruled}{labelfont=normalfont,labelsep=colon,strut=off} 
\usepackage{algorithm}
\usepackage{algorithmic}

\usepackage{newfloat}
\usepackage{listings}
\floatstyle{ruled}
\newfloat{listing}{tb}{lst}{}
\floatname{listing}{Listing}
\title{Panini-Net: GAN Prior Based Degradation-Aware Feature Interpolation \\for Face Restoration}
\author{
    Yinhuai~Wang\textsuperscript{\rm 1},
    Yujie~Hu\textsuperscript{\rm 1},
    Jian~Zhang\textsuperscript{\rm 1,2}}
\affiliations{
    \textsuperscript{\rm 1}Peking University Shenzhen Graduate School, China\\
    \textsuperscript{\rm 2}Peng Cheng Laboratory, China

    \{yinhuai; hhuyujie\}@stu.pku.edu.cn, zhangjian.sz@pku.edu.cn
}

\usepackage{bibentry}

\begin{document}
\maketitle

\begin{abstract}
Emerging high-quality face restoration (FR) methods often utilize pre-trained GAN models (\textit{i.e.}, StyleGAN2) as GAN Prior. However, these methods usually struggle to balance realness and fidelity when facing various degradation levels. Besides, there is still a noticeable visual quality gap compared with pre-trained GAN models. In this paper, we propose a novel GAN Prior based degradation-aware feature interpolation network, dubbed Panini-Net, for FR tasks by explicitly learning the abstract representations to distinguish various degradations. Specifically, an unsupervised degradation representation learning (UDRL) strategy is first developed to extract degradation representations (DR) of the input degraded images. Then, a degradation-aware feature interpolation (DAFI) module is proposed to dynamically fuse the two types of informative features (\textit{i.e.}, features from input images and features from GAN Prior) with flexible adaption to various degradations based on DR. Ablation studies reveal the working mechanism of DAFI and its potential for editable FR. Extensive experiments demonstrate that our Panini-Net achieves state-of-the-art performance for multi-degradation face restoration and face super-resolution. The source code is available at https://github.com/jianzhangcs/panini.\let\thefootnote\relax\footnotetext{This work was supported in part by Shenzhen Fundamental Research Program (No.GXWD20201231165807007-20200807164903001). \textit{ Corresponding author: Jian~Zhang}.}
\end{abstract}

\section{Introduction}

Face restoration (FR) is typically an ill-posed image inverse problem, especially under high-degradation or multi-degradation (\textit{e.g.}, downsampling, noise, blur, and compression) cases. Traditional deep network-based methods often utilize a single model for end-to-end training \cite{esrgan, hifacegan}, which can well grasp the overall structure but lack the richness of details.

Due to the excellent performance of GAN models in recent years \cite{stylegan,stylegan2,pggan}, some methods \cite{pulse, psp} begin to use pre-trained GAN models as GAN Prior for FR tasks. These methods take advantage of the rich details implicitly encapsulated in GAN Prior by encoding the degraded face image into the latent space of pre-trained GAN. Although high visual quality is achieved, due to the low dimension of latent space and its poor spatial expression capability, these methods are often unable to fully capture the facial structures of the degraded face image, which usually manifests as identity inconsistency.

To further capture the facial structural information of the degraded face image while preserving the realness contributed by GAN Prior, some methods \cite{glean, gpen, gfpgan} not only encode the degraded face image into the latent space but also fuse external features (\textit{e.g.}, features extracted from the degraded face image) with the GAN Prior features. These methods achieve significant improvements in identity consistency than previous GAN Prior based FR methods. However, they don't provide an explicit design for degradation-aware feature fusion and consequently result in inadequate robustness in visual quality when facing different degradation levels.

Inspired by recent progress in contrastive learning \cite{moco, moco2, simclr, simclr2} and visual attention \cite{senet, cbam, sknet, dasr}, an unsupervised degradation representation learning (UDRL) strategy has been proposed in this paper to pre-train a degradation representation encoder (DRE). DRE extracts the degradation representations (DR) of input degraded images, regarded as a global condition to guide the restoration process. In addition, we propose a novel degradation-aware feature interpolation (DAFI) module, which can dynamically fuse the features (\textit{i.e.}, features of GAN Prior and features extracted from the degraded face image) according to DR. 

Further, we propose a novel network to integrate these designs for FR tasks. As the idea of selecting and fusing different sources of features is interestingly similar to the way of making a panini, we dubbed our network as Panini-Net. For instance, the degraded face image usually contains abundant valid information when suffering from mild degradation. Then, our Panini-Net will increase the fusion proportion of degraded image features while reducing the fusion proportion of GAN Prior features. The degraded face image usually lacks valid information when the degradation is severe. Then, Panini-Net will decrease the fusion proportion of degraded image features while increasing the proportion of GAN Prior features. Besides, the proposed DAFI shows some interesting attributes. In our experiments and ablation studies, DAFI performs better in details compared with normal convolutional feature fusion methods, even if DAFI possesses much fewer parameters. DAFI also shows flexible editability in FR tasks, and this attribute enables Panini-Net to generate multiple high-quality restored images.

\textbf{The main contributions of this paper are summarized as follows:}
\begin{itemize}
\item We propose a novel GAN Prior based degradation-aware feature interpolation network for FR, dubbed Panini-Net. It provides a robust solution to balance realness and fidelity considering various degradation levels. 

\item We propose an unsupervised degradation representation learning strategy to extract discriminative degradation representations for degraded images, which explicitly serve as a global condition for dynamic fusion. 

\item  We propose a novel degradation-aware feature interpolation (DAFI) module, which can dynamically fuse the two types of informative features from input images and GAN Prior based on different degradation representations. We also experimentally show the efficiency and editability of DAFI.

\item Experiments on two typical FR tasks, \textit{i.e.}, multi-degradation face restoration, 16$\times$ super-resolution, show that Panini-Net achieves state-of-the-art results.
\end{itemize}

\section{Related Work}
\paragraph{Face Restoration.} The use of deep neural networks (DNNs) in FR has made great progress in recent years \cite{srcnn, vdsr, rcan, rdn, wavelet, xiaoming, lin2018multi, lin2020learning}. General DNNs based FR methods use paired datasets for end-to-end training. However, they struggle to generate high-quality images. With the development of Generative Adversarial Network (GAN) \cite{gan, cgan}, some image restoration methods begin to introduce adversarial training strategies to improve the visual quality of repaired images \cite{esrgan, xinyu, hifacegan, ziyu, menglei}. Those methods can restore the overall structure of the image very well but are still inferior to the best generative models \cite{stylegan, stylegan2} in visual quality. The human face has regular structures and details, making it possible to use strong prior information for FR. Commonly used priors include parsing maps \cite{deepsee, xinyu2, fsrnet, psfr, song2019geometry}, landmarks \cite{xiaobin, fsrnet, song2019geometry}, reference images \cite{xiaoming, xiaoming3}, and GAN Prior \cite{glean, gpen, gfpgan}. Recent GAN Prior based FR methods achieve unprecedented results.

\paragraph{GAN Prior.} In recent years, GANs represented by StyleGAN \cite{stylegan}, which can generate high-resolution images, have inspired a lot of GAN Prior based image editing works \cite{pulse, psp, sam, styleflow, timetravel, image2stylegan, image2stylegan++, iddisentanglement, stylerig, sean, styleclip}. These methods often implement image editing by GAN Inversion. Specifically, the input image is first embedded into the latent space of StyleGAN as a latent code. Then the latent code is controlled to achieve image editing, such as super-resolution, inpainting, attribute transformation, \textit{etc}. In PULSE \cite{pulse}, the authors optimize the latent code of StyleGAN to solve image super-resolution tasks. In PSP \cite{psp}, the authors train an encoder to directly predict the latent code instead of iterative optimizing it, thus accelerating the inference time. However, the common flaw for latent code editing is that it can not capture the spatial structures well. Therefore, the identities are usually inconsistent between the input image and the edited one. Even so, the advantages of latent code editing are evident, and the resulting images are of high quality, with realistic details. 

Considering the image realness of StyleGAN and the image structural fidelity of the general DNN based methods, recent methods \cite{glean, gpen, gfpgan} try to combine those two merits for FR by fusing external structural features into the intermediate features of pre-trained StyleGAN. GPEN \cite{gpen} utilizes a pre-trained StyleGAN2 generator as the Decoder. Given a degraded image as input, the Encoder predicts a latent code of StyleGAN2 and extracts intermediate features as the input noise of StyleGAN2. GLEAN \cite{glean} utilizes StyleGAN2 generator as the latent bank and uses an encoder-latent bank-decoder framework for super-resolution. The features connected between each part are fused in a concatenation-convolution way. GFP-GAN \cite{gfpgan} uses a degradation removal module to predict latent codes and extracts the intermediate features to modulate the StyleGAN features. The degradation removal module is supervised during training to improve the quality of extracted features. In addition, the eyes and mouth are separately trained by GAN to improve quality. These methods achieve significant improvements in identity consistency than previous GAN Prior based FR methods, but they don't have an explicit design for dynamic feature fusion considering different degradation levels, thus showing insufficient robustness in image visual quality when facing severe degradations. 

In Image2StyleGAN++ \cite{image2stylegan++}, the authors find that editing the ``activation tensors" of StyleGAN could achieve more precise spatial-wise editing. Luo \textit{et al.} carry out style-mix on the lower layers of latent codes $\mathbf{w}^{+}$ to migrate the target identity \cite{timetravel}. Barbershop \cite{barbershop} replaces the output ``activations" of a specific layer in StyleGAN2 to achieve better identity migration. These works reveal that the bottom layers of StyleGAN have greater influences on the coarse structures, while the higher layers mainly contribute to the details. These characteristics of StyleGAN inspire us to solve FR tasks by editing the bottom features of StyleGAN to achieve identity consistency while keeping the higher layers untouched to preserve the textures and facial details.

\begin{figure*}[t]
  \centering
  \includegraphics[width=1\linewidth]{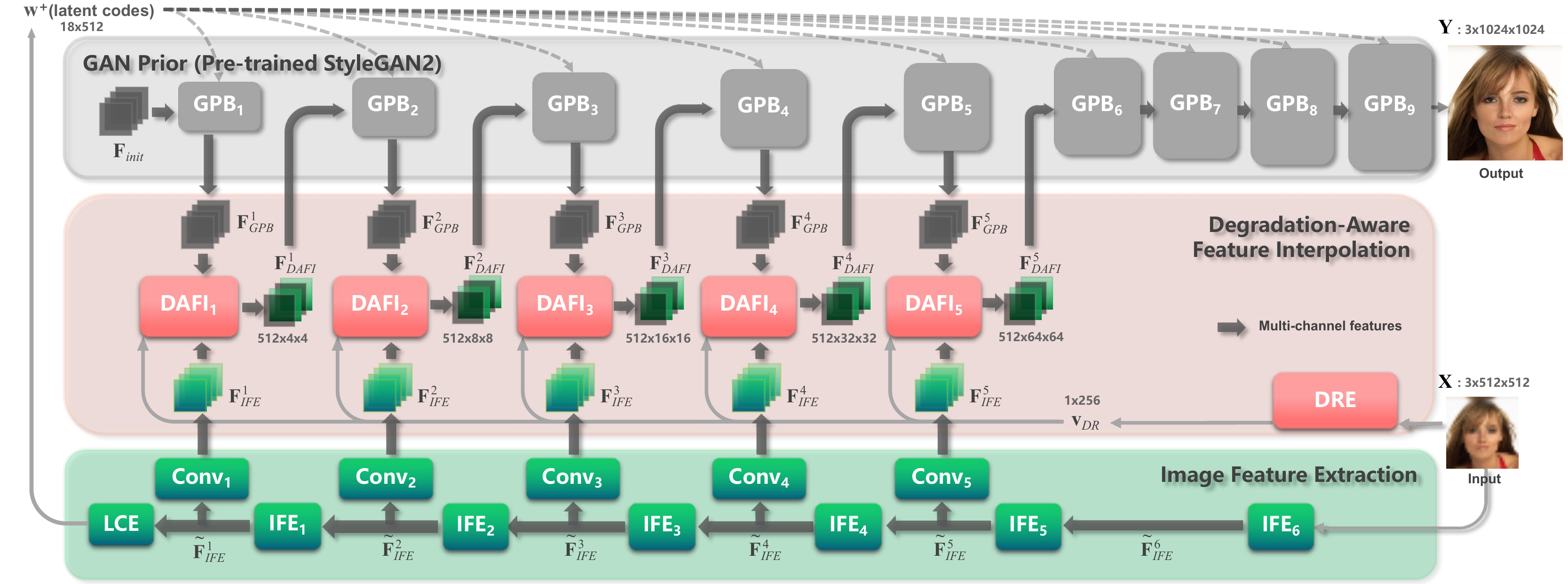}
  \caption{Overview of Panini-Net. It consists of an image feature extraction module, a degradation-aware feature interpolation (DAFI) module, and a pre-trained StyleGAN2 as GAN Prior module (GPM). Given a degraded face image $\mathbf{X}$ as the input, the image feature extraction module extracts features $\mathbf{F}_{IFE}^{i}, i\in\{1, ..., 5\}$, and predicts latent codes $\mathbf{w}^{+}\in \mathbb{R}^{18\times 512}$. The latent codes $\mathbf{w}^{+}$ can coarsely fetch a similar high-quality face from GPM. Then, 5 DAFI blocks (denoted as DAFI$_{i}$) are used to progressively interpolate $\mathbf{F}_{IFE}^{i}$ into $\mathbf{F}_{GPB}^{i}$ to incorporate the valid structural information of the degraded face image. A pre-trained degradation representations encoder (DRE) encodes the degradation representations as a vector $\mathbf{v}_{DR}$, which can be regarded as a global condition to guide DAFI blocks for restoration.}
\label{panini-net} 
\end{figure*}

\paragraph{Contrastive Learning.} Recently, contrastive learning \cite{moco, simclr, simclr2, moco2} has made great strides in unsupervised representational learning. Some methods begin to use contrastive learning for image generation tasks \cite{cut, divco}. In DASR \cite{dasr}, the authors use the MoCo framework to pre-train a degradation encoder to learn degradation representations, then use degradation representations to guide the network for super-resolution tasks. In the pre-training of the degradation encoder, two patches are randomly selected from the same degraded image, and one patch is taken as a positive example for the other. These two patches share the same degradation functions, while the contents may also be similar. Therefore, the degradation encoder may learn not only the degradation representations but also the content representations. Moreover, for FR tasks, a portrait photo is often shot with a large aperture. The accompanying depth of field (DOF) may blur the background. If the patch of DASR is happened to locate in the background, the degradation encoder may be hard to differentiate whether the background blur is caused by DOF or by degradation. Therefore, we design an unsupervised degradation representation learning (UDRL) strategy to focus on learning the overall degradation representations of the degraded face image and encourage learning the degradation rather than the contents.

\paragraph{Visual Attention.} Visual attention is usually explored to improve the performance of CNN, which not only tells where to focus but also improves the representation of interests. In \cite{senet}, the authors propose the Squeeze-and-Excitation (SE) block, which adaptively adjusts channel-wise feature weights by explicitly modeling interdependencies between channels. In \cite{cbam}, the proposed Convolutional Block Attention Module (CBAM) explicitly incorporates both channel-wise and spatial-wise attention. In \cite{sknet}, the Selective Kernel (SK) unit adaptively adjusts the receptive field size based on the region of interest. These methods provide robust baselines for universal visual tasks. However, they all extract attention from local features which need to be processed, while in FR models, these features may hardly contain valid degradation information. Besides, they can only reinforce features from a single source, while our urgent demand is to dynamically fuse two sources of features (\textit{i.e.}, features from GAN Prior and features extracted from degraded face image). To overcome these deficiencies, our proposed DAFI module extracts the global condition directly from the degraded face image and enforces adaptive feature fusion for features from distinct sources. 

\section{Proposed Panini-Net}
An overview of Panini-Net is depicted in Fig.~\ref{panini-net}. Specifically, Panini-Net consists of an image feature extraction module, a pre-trained face GAN model (\textit{e.g.}, StyleGAN2) as the GAN Prior module, and a degradation-aware feature interpolation module.  Given a degraded image $\mathbf{X}$, its degradation representation (DR) is encoded by the pre-trained degradation representation encoder (DRE) as $\mathbf{v}_{DR}$. The image feature extraction module extracts features $\mathbf{F}_{IFE}$ from the degraded face image and generates latent codes $\mathbf{w}^{+}$. $\mathbf{w}^{+}$ can coarsely fetch a similar high-quality face from GAN Prior module. Then the degradation-aware feature interpolation module progressively interpolates features $\mathbf{F}_{IFE}$ into GAN Prior features to rectify the facial structures of the fetched similar face, and finally, obtain a high-quality face image $\mathbf{Y}$ with realness and identity consistency.

\begin{figure}[t]
  \centering
   \vspace{-2pt}
  \includegraphics[width=1\linewidth]{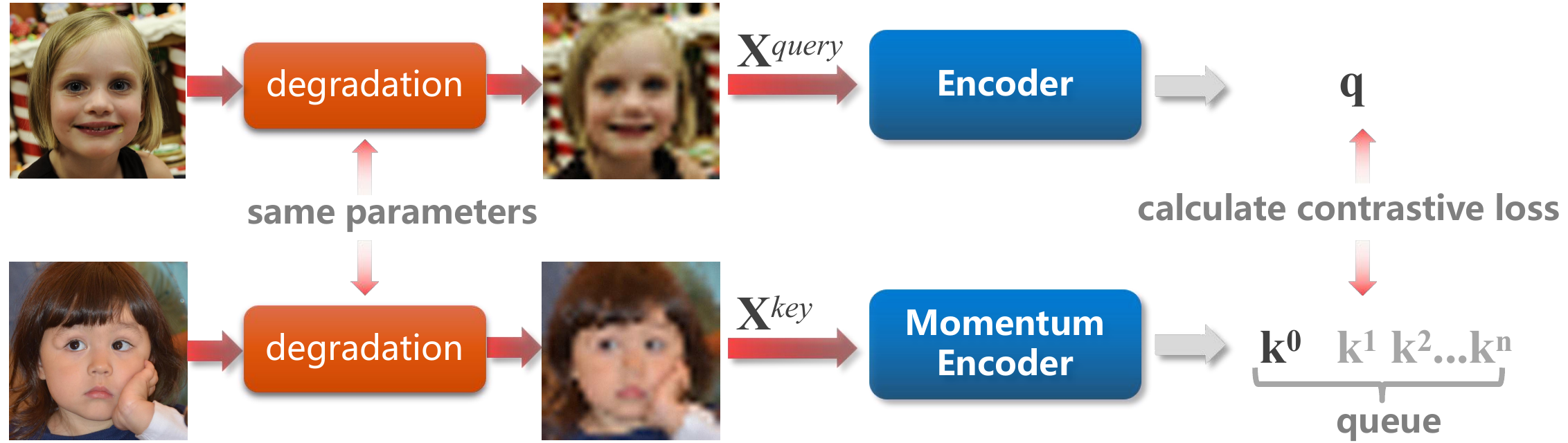}
  \caption{Overview of the unsupervised degradation representation learning strategy for degradation representation encoder (DRE). For each iteration, we randomly generate a set of new degradation parameters, and operate them on two different new HQ images to generate positive example pairs. Let the history images in queue be negative examples, to encourage learning the degradation rather than the contents.
  }
 \vspace{-8pt}
\label{DP} 
\end{figure}

\subsection{Image Feature Extraction Module} 
As shown in Fig.~\ref{panini-net}, the image feature extraction module is designed to extract features $\mathbf{F}_{IFE}$ from the degraded face image and generate latent codes $\mathbf{w}^{+}$. Given an input image $\mathbf{X}$, we use image feature extractor (IFE) to extract preliminary features $\mathbf{\tilde{F}}_{IFE}^{i}\in\mathbb{R}^{C_i\times H_i\times W_i}$ progressively:
\begin{equation}
    \mathbf{\tilde{F}}_{IFE}^{i} =  \begin{cases}\mathcal{H}_{IFE}^{i}(\mathbf{X}), & i = 6;\\\mathcal{H}_{IFE}^{i}(\mathbf{\tilde{F}}_{IFE}^{i+1}), & 1\le i<6,\end{cases}
\end{equation}
where $\mathcal{H}_{IFE}^{6}(\cdot)$ is a dense block, while $\mathcal{H}_{IFE}^{i}(\cdot)$, $i\in\{1, ..., 5\}$, is just a convolution layer.  

In order to avoid coupling of adjacent features, we add additional convolution branch to further extract features $\mathbf{F}_{IFE}^{i}\in\mathbb{R}^{C_i\times H_i\times W_i}$  as the final features for fusion:
\begin{equation}
    \mathbf{F}_{IFE}^{i} =  \mathcal{H}_{Conv}^{i}(\mathbf{\tilde{F}}_{IFE}^{i}),\quad i\in\{1, ..., 5\},
\end{equation}
where $\mathcal{H}_{Conv}^{i}(\cdot)$ is just a convolution layer, which has been experimentally verified useful. 

Finally, a latent code encoder (LCE), which is composed of a convolution layer and a fully-connected layer, is used to predict the latent codes $\mathbf{w}^{+}\in\mathbb{R}^{18\times512}$, expressed  as
\begin{equation}
    \mathbf{w}^{+} =  \mathcal{H}_{LCE}(\mathbf{\tilde{F}}_{IFE}^{1}).
\end{equation}

Note that in our Panini-Net, as illustrated in Fig.~\ref{panini-net}, $\mathbf{\tilde{F}}_{IFE}^{i}$, $ \mathbf{F}_{IFE}^{i}$, $\mathbf{F}_{GPB}^{i}$, $\mathbf{F}_{DAFI}^{i}$ have the same size of $C_i$$\times$$H_i$$\times W_i$.

\subsection{GAN Prior Module} 
A pre-trained StyleGAN2 generator \cite{stylegan2} is utilized as the GAN Prior module (GPM) in our Panini-Net. To be concrete, as shown in Fig.~\ref{panini-net}, GPM starts with a learned constant features $\mathbf{F}_{init}$, then progressively generates the result by passing $\mathbf{F}_{init}$ through a series of GAN Prior blocks (GPB$_{i}$, $i\in\{1, ..., 5\}$). Each GPB$_{i}$ includes an up-sample operation (except for $\text{GPB}_{1}$) and outputs the feature $\mathbf{F}_{GPB}^{i}$. The latent codes $\mathbf{w}^{+}$ can coarsely fetch a similar high-quality face from GPM. The features $\mathbf{F}_{GPB}^{i}$, $i\in\{1, ..., 5\}$, are selected for dynamic fusion with our proposed degradation-aware feature interpolation module to rectify the facial structures. To preserve the delicate facial details, we leave the rest part of GPM untouched. It is worth noting that we omit details in StyleGAN2 (or GPM, GPB) for simplicity. For more details about StyleGAN2 please refer to \cite{stylegan, stylegan2}.

\subsection{Degradation-Aware Feature Interpolation Module}
The degradation-aware feature interpolation (DAFI) module is comprised of a degradation representation encoder (DRE) and several DAFI blocks (denoted as DAFI$_{i}$, $i\in\{1, ..., 5\}$).

The DRE is designed to extract the degradation representations (DR) of the degraded face images. We adopt an unsupervised degradation representation learning (UDRL) strategy to pre-train DRE. Specifically, as illustrated in Fig.~\ref{DP}, two degraded images $\mathbf{x}^{query}$ and $\mathbf{x}^{key}$ are obtained by applying the same degradation function on two different images. Then we take $\mathbf{x}^{key}$ as the positive example of $\mathbf{x}^{query}$ and use MoCo \cite{moco2} framework to conduct contrastive training. We take $\mathbf{x}^{query}$ as the input for Encoder to generate vector $\mathbf{q}$, and take $\mathbf{x}^{key}$ as the input for the Momentum Encoder to generate vector $\mathbf{k}^{0}$. The history generated by Momentum Encoder can form a queue $\mathbf{k}^{0},\mathbf{k}^{1}, ... , \mathbf{k}^{n}$. For each $\mathbf{q}$, as $\mathbf{k}^{0}$ is generated from $\mathbf{x}^{key}$ which shares the same degradation mode with $\mathbf{x}^{query}$, vector $\mathbf{k}^{0}$ should be similar to $\mathbf{q}$, while the rest vectors in the queue should be different from $\mathbf{q}$. An InfoNCE loss is used as training objective to encourage the encoding of $\mathbf{q}$ to approach $\mathbf{k}^{0}$ and stay away from $\mathbf{k}^{1},\mathbf{k}^{2}, ... , \mathbf{k}^{n}$, which is formulated as:
\begin{equation}
    \mathcal{L}_{DR}=\sum_{i=1}^{B}-\log\frac{\exp(\mathbf{q}_{i}\cdot \mathbf{k}_{i}^{0}/\tau)}{\sum_{j=1}^{n}\exp(\mathbf{q}_{i}\cdot \mathbf{k}_{i}^{j}/\tau)}.    
\end{equation}

We first pre-train the DRE under MoCo framework with our UDRL strategy, then fine-tune Panini-Net with fixed DRE. Given a degraded image $\mathbf{X}$, its degradation representation (DR) is efficiently encoded as a vector $\mathbf{v}_{DR}\in\mathbb{R}^{1\times256}$:
\begin{equation}
    \mathbf{v}_{DR} =  \mathcal{H}_{DRE}(\mathbf{X}).    
\end{equation}

\begin{figure}[t]
  \centering
   \vspace{-2pt}
  \includegraphics[width=1\linewidth]{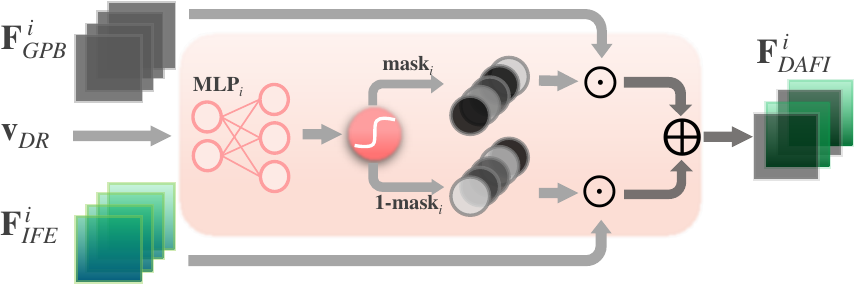}
  \caption{Overview of the degradation-aware feature interpolation (DAFI) block. $\mathbf{v}_{DR}$ serves a condition to generate an adaptive degradation-aware channel-wise $\mathbf{mask}$ through an MLP followed by Softmax. Then, the $\mathbf{mask}$ is used for dynamic interpolation.}
\label{TIMO} 
 \vspace{-8pt}
\end{figure}

For each feature $\mathbf{F}_{GPB}^{i}$ generated by $\text{GPB}_{i}$, $i\in\{1, ..., 5\}$, a dedicated DAFI block is applied to incorporate the valid facial structural information of its counterpart $\mathbf{F}_{IFE}^{i}$. 

As Fig.~\ref{TIMO} shows, given $\mathbf{v}_{DR}$ as the global condition, each DAFI block first applies a dedicated MLP followed by Softmax operation to generate an adaptive degradation-aware channel-wise $\mathbf{mask}_i\in \mathbb{R}^{1\times C_i}$, $i\in\{1, ..., 5\}$, that is  
\begin{equation}
    (\mathbf{mask}_{i},\mathbf{1-mask}_{i}) ={\text{Softmax}}(\mathcal{H}_{MLP}^{i}(\mathbf{v}_{DR})).
\end{equation}

Obviously, the $\mathbf{mask}_i$ is a vector whose dimension is equal to the channel number of features to be interpolated, and each $\mathbf{mask}$ element represents an interpolation weight of a certain channel. Then  the degradation-aware interpolation between image feature and GAN prior is formulated as:
\begin{equation}
    \mathbf{F}_{DAFI}^{i} =\mathbf{F}_{GPB}^{i}\odot\mathbf{mask}_{i}+\mathbf{F}_{IFE}^{i}\odot(1-\mathbf{mask}_{i}),
\label{eq7} 
\end{equation}
where $\odot$ denotes channel-wise multiplication. 

Next, we replace $\mathbf{F}_{GPB}^{i}$ with $\mathbf{F}_{DAFI}^{i}$ as the input for the $(i+1)$-th GAN prior block  $\text{GPB}_{i+1}$, $i\in\{1, ..., 5\}$, \textit{i.e.},
\begin{equation}
    \mathbf{F}_{GPB}^{i} =  \begin{cases}\mathcal{H}_{GPB}^{i}(\mathbf{F}_{init}), & i = 1\\\mathcal{H}_{GPB}^{i}(\mathbf{F}^{i-1}_{DAFI}), & 1<i<7\\\mathcal{H}_{GPB}^{i}(\mathbf{F}^{i-1}_{GPB}). & 7 \leq i<9\end{cases}
\end{equation}

The final result $\mathbf{Y}$ is generated by the last block $\text{GPB}_{9}$ as
\begin{equation}
    \mathbf{Y} =  \mathcal{H}_{GPB}^{9}(\mathbf{F}^{8}_{GPB}).
\end{equation}

\subsection{Training and Loss}
We first pre-train the DRE by UDRL strategy, then load the pre-trained StyleGAN2 \cite{stylegan2} generator as our GPM. Finally, we fine-tune the whole Panini-Net. To provide stable degradation representations, we fix the parameters of DRE during fine-tuning. The rest parameters of Panini-Net are learnable. We use standard L1 loss, VGG perceptual loss \cite{perceptualloss}, and vanilla adversarial loss \cite{gan} as objectives for fine-tuning. The pre-trained discriminator of StyleGAN2 is also used for adversarial training. Details about the training and loss functions can be found in the appendix.

\section{Experimental Results}
To verify the utility of our Panini-Net on different FR tasks, we conduct extensive experiments on two typical FR tasks: multi-degradation face restoration and face super-resolution.

\begin{figure}[t]
  \centering
   \vspace{-2pt}
  \includegraphics[width=1\linewidth]{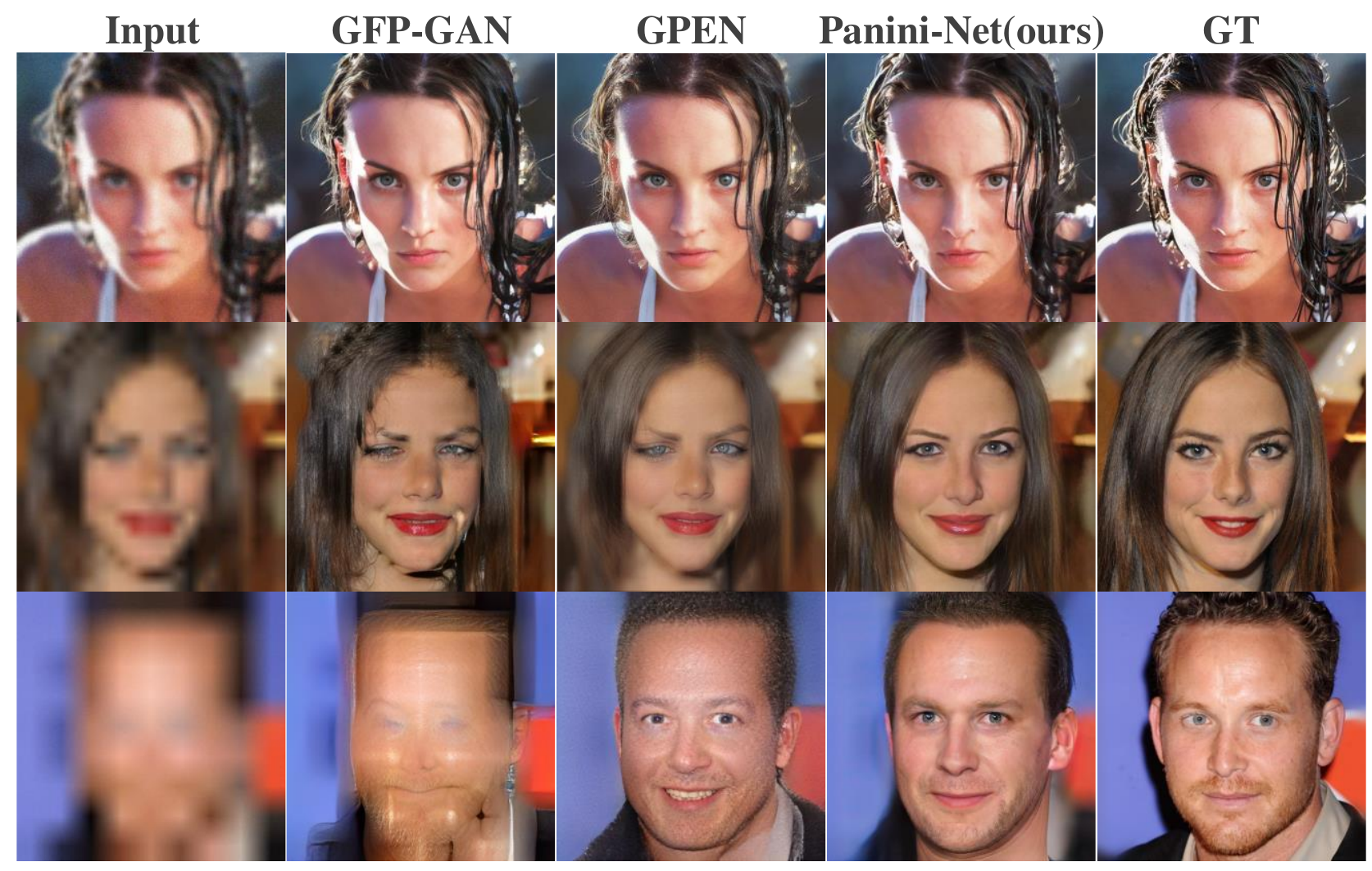}
  \caption{Experiment with multi-degradation face restoration. Given input images with different degradation levels, GFP-GAN performs poorly when degradation is severe, and GPEN struggle to generate realistic details, while our Panini-Net can achieve both highest realness and identity consistency. \textbf{Zoom in for best view}.}
\label{MFRexp} 
 \vspace{-8pt}
\end{figure}

\begin{table}[t]
    \centering
        \caption{Quantitative comparison on multi-degradation face restoration. Our Panini-Net achieves the highest PSNR and competitive LPIPS. A significant gain in FID is also obtained, which statistically reflects the improvement of image realness contributed by our method.}
    \begin{tabular}{lccc}
        \hline
            \rule{0pt}{10pt}{\textbf{Method}} & PSNR↑ & FID↓ & LPIPS↓ \\
        \hline
            \rule{0pt}{10pt}{GFP-GAN (CVPR'21)} & 17.22 & 34.61 & \textbf{0.4150}\\
            \rule{0pt}{10pt}{GPEN (CVPR'21)} & 17.91 & 32.03& 0.4355\\
            \rule{0pt}{10pt}{Panini-Net (ours)} &\textbf{18.01}&\textbf{24.66}&0.4470\\
         \hline
    \end{tabular}
    \label{tb_mfr}
\end{table}

\subsection{Face Restoration with Multiple Degradations} 
We train our Panini-Net on the FFHQ dataset \cite{stylegan}, which contains 70000 high-quality (HQ) face images. We follow the practice in \cite{gpen, gfpgan} that applies a widely used degradation function to synthesize degraded low-quality (LQ) images:
\begin{equation}
    \mathbf{X} =[(\mathbf{Y}\circledast\mathbf{k}_{\sigma})\mathbf{\downarrow}_{r} + \mathbf{N}_{\delta}]_{\mathbf{JPEG}_{q}}.
\label{deg} 
\end{equation}

Specifically, HQ image $\mathbf{Y}$ is first convolved by Gaussian blur kernel $\mathbf{k}$ followed by a downsampling operation (with a symmetric upsampling to keep the size invariant), then an additive noise is applied on it. Finally, the image is compressed by JPEG operation with quality factor $q$. For each HQ image $\mathbf{Y}$, the blur kernel size $\sigma$ is randomly selected. ${r}$, ${\delta}$, $q$ are randomly chosen from [10:200], [0:25], [5:50] respectively. By this means, we generate HQ-LQ image pairs for training. LQ and HQ images are of size $512$$\times$$512$ and $1024$$\times$$1024$, respectively. We adopt Adam optimizer and Cosine Annealing Scheme to fine-tune Panini-Net, with batch size being 8 and iterations number being 600K.

We randomly choose 1000 HQ face images in CelebA-HQ \cite{pggan} dataset as ground truth (GT), then use the above degradation function Eq.~\eqref{deg} to process these images as input. We compare Panini-Net with recent state-of-the-art GAN Prior based FR methods, including official pre-trained GFP-GAN \cite{gfpgan} and GPEN \cite{gpen} models. It is worth noting that these competing methods all use StyleGAN2 as GAN Prior and share the same degradation formulation (may be different in parameters) for synthesizing LQ images. We use PSNR, FID \cite{fid}, LPIPS \cite{lpips} as the metrics for quantitative comparison. Subjective and visual results are shown in Table~\ref{tb_mfr} and  Fig.~\ref{MFRexp}, which clearly shows the evident advantage of Panini-Net in the balance between image realness and identity consistency. The images restored by Panini-Net are comparable with the ground truth in visual quality even when the degradation is severe.

\begin{figure}[t]
  \centering
   \vspace{-2pt}
  \includegraphics[width=1\linewidth]{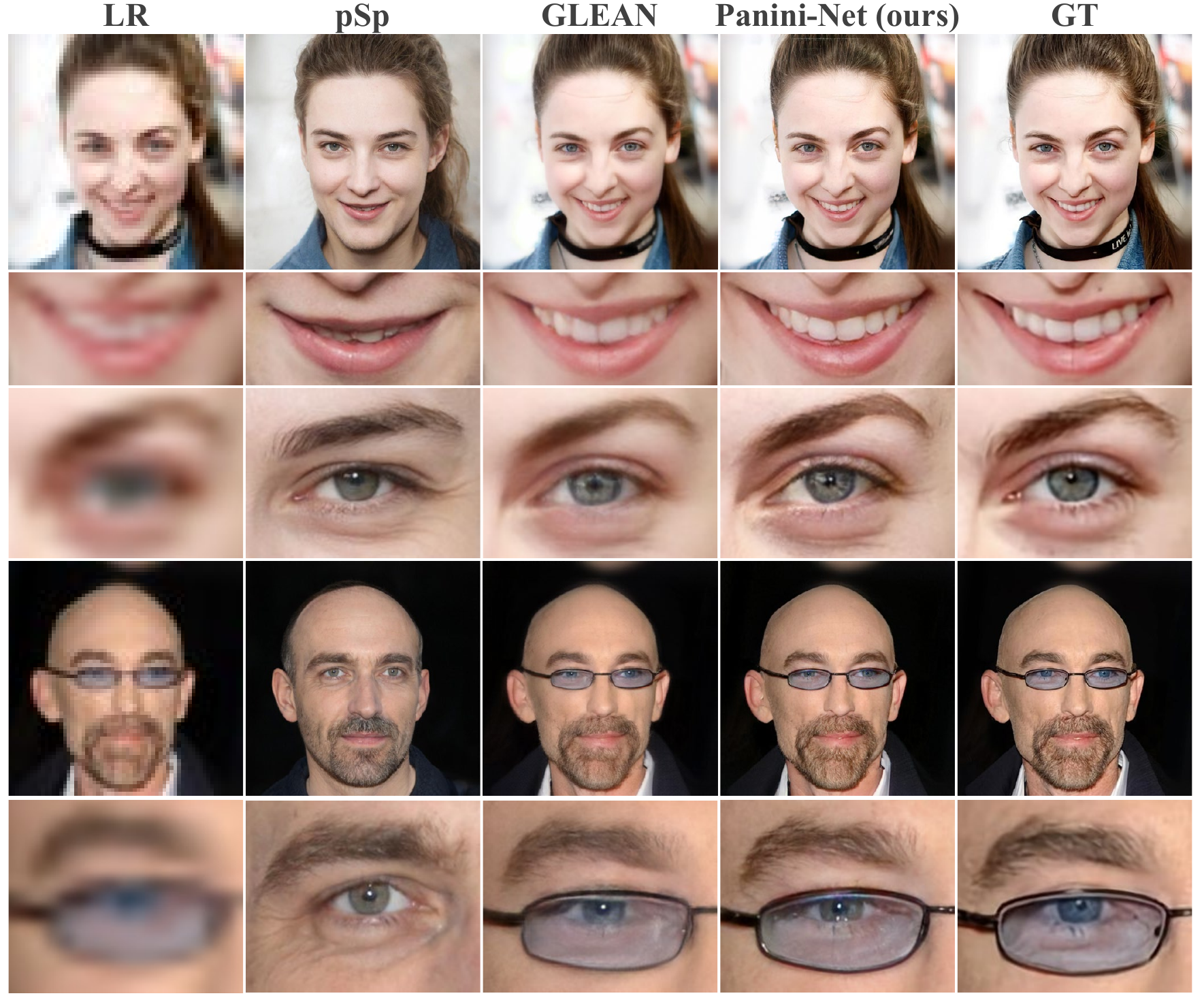}
  \caption{Experiment with $16\times$ SR. Despite pSp can achieve high visual quality, it performs poorly in maintaining the identity of the input face image. GLEAN can achieve good results, while Panini-Net provides comparable even superior performance, especially in details (\textit{e.g.}, eyes and tooth). Since Panini-Net has much fewer parameters than GLEAN, and they share similar training frameworks, we attribute its superiority to the proposed DAFI. \textbf{Zoom in for best view}.}
\label{SRtest} 
 \vspace{-8pt}
\end{figure}

\begin{table}[t]
    \centering
    \begin{tabular}{lccc}

      \hline
         \rule{0pt}{10pt}{\textbf{Method}} & PSNR↑ & FID↓ & LPIPS↓ \\
         \hline
         \rule{0pt}{10pt}{pSp (CVPR'21)} &12.90&47.65&0.6529\\
         \rule{0pt}{10pt}{GLEAN (CVPR'21)} &\textbf{21.66}&19.76&0.4013\\
         \rule{0pt}{10pt}{Panini-Net (ours)} &21.19&\textbf{16.77}&\textbf{0.3886}\\
         \hline
    \end{tabular}
    \label{tb_16xsr}
\caption{Quantitative comparison on $16\times$ face SR. Our Panini-Net outperforms other methods in FID and LPIPS, and achieves competitive PSNR. As is shown in Fig. 5, the quality of our results
outperforms the GT sometimes, but metrics
like PSNR can not well reflect that, as they only calculates
the consistency with the GT, without any consideration for
other reasonable multiple solutions.}    
\end{table}

\subsection{Face Super-Resolution}
Downsampling with a fixed ratio can be considered as a constant degradation. Thus we don't need a DRE to extract the degradation representations. Specifically, we simplified Panini-Net for the $16\times$ SR task: (1) We remove DRE and represent $\mathbf{v}_{DR}$ with a learnable constant vector. (2) Some relevant convolutions are adjusted to fit the new size of the input image. In this experiment, we use the FFHQ dataset as the GT and use $16\times$ bilinear interpolation as a downsampling operation to generate low-resolution (LR) images, thus forming GT-LR pairs for training. The size of LR images is $64$$\times$$64$, and $1024$$\times$$1024$ for GT, corresponding to the input and output size of the modified Panini-Net. We use the same way to generate a test dataset from CelebA-HQ and compare Panini-Net with recent SOTA GAN Prior based SR methods on our test dataset, including pSp \cite{psp}, and GLEAN \cite{glean}, which also use StyleGAN2 as GAN Prior. Results are shown in Fig. \ref{SRtest} and Table \ref{tb_16xsr}. Although with fewer parameters, our Panini-Net is still competitive even outperforms these methods considering its superiority in the balance between realness and fidelity.

\section{Ablation Studies}

\subsection{\textbf{A}: Comparison with Different Feature Fusion}
\begin{figure}[t]
  \centering
  \vspace{-2pt}
  \includegraphics[width=\linewidth]{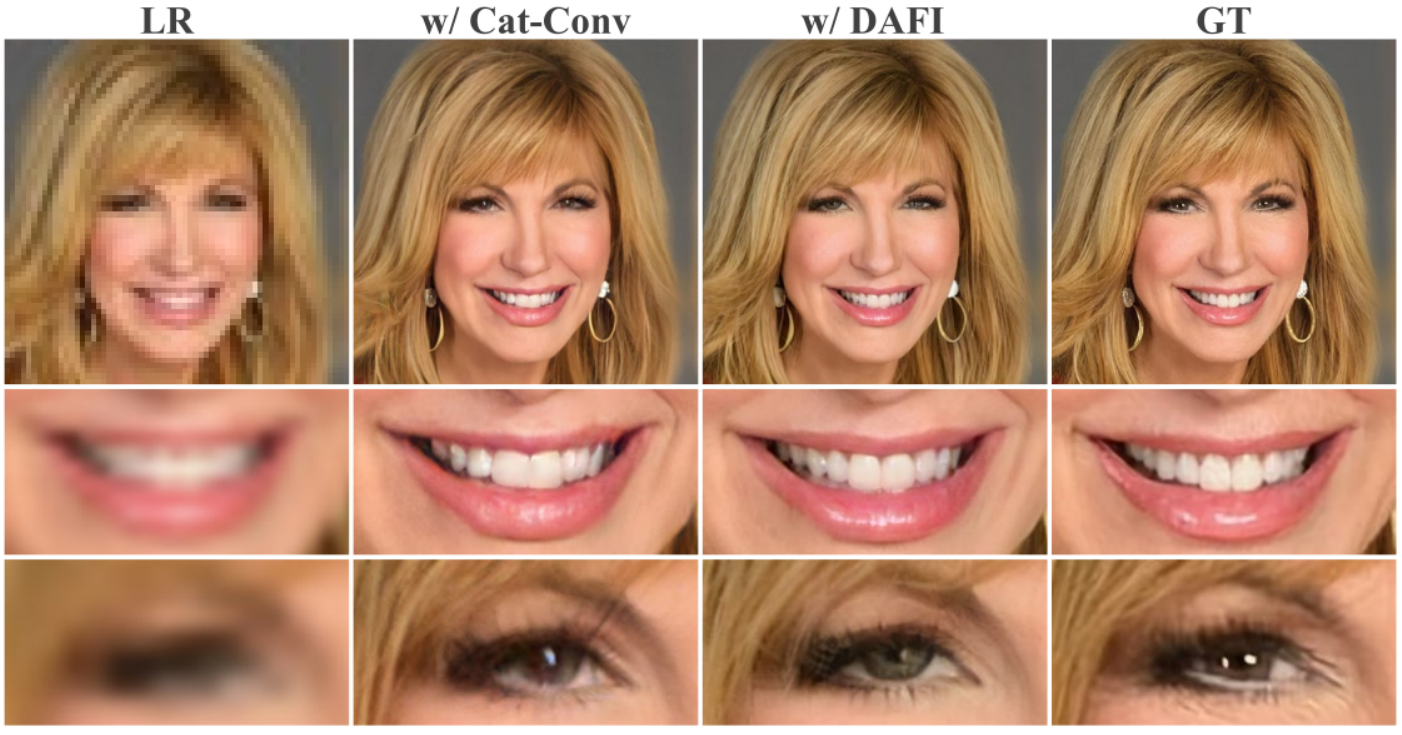}
  \caption{Ablation experiment \textbf{A}. Panini-Net with Cat-Conv is a modified version of Panini-Net, which removes DRE and replaces DAFI with concatenation followed by convolution operations (similar feature fusion operation in GLEAN). Cat-Conv costs more parameters and undermines the visual quality, especially in details (\textit{e.g.}, tooth and eyes). We argue that's because the interpolation operation in DAFI can better preserve the details encapsulated in GAN Prior features, and the global condition guidance can help DAFI better handle feature fusion. \textbf{Zoom in for best view}.}
\label{ablation A} 
\vspace{-8pt}
\end{figure}

GAN Prior based FR methods \cite{glean, gfpgan, gpen} usually achieve identity consistency by fusing external features into features of the GAN Prior model. Concatenation followed by convolution (denoted as Cat-Conv in this paper) is a typical feature fusion method that is also adopted in GLEAN \cite{glean}. Given two informative features from different sources, it first concatenates them by channel, then uses a convolution layer to decrease the channel number. Here we simply replace DAFI in Panini-Net with Cat-Conv. To exclude the effect of DRE, We train Panini-Net with Cat-Conv on the $16\times$ SR task (without DRE) with the same training settings as Panini-Net. Results are shown in Fig. \ref{ablation A} and Table \ref{tb_ablation a}. 

\begin{table}[t]
    \centering
        \caption{Quantitative comparison on different feature fusion. Concatenation will double the feature channels. Considering features with 512 channels, Cat-Conv will be costly in parameters. Instead, DAFI uses interpolation for feature fusion with fewer parameters and performs better in visual quality and quantitative metrics.}
    \begin{tabular}{lcc}

      \hline
         \rule{0pt}{10pt}{\textbf{Method}} & PSNR↑ & FID↓\\
         \hline
         \rule{0pt}{10pt}{Panini-Net with Cat-Conv} &21.17&18.41\\
         \rule{0pt}{10pt}{Panini-Net with DAFI (ours)} &\textbf{21.19}&\textbf{16.77}\\
         \hline
    \end{tabular}
    \label{tb_ablation a}
\end{table}

Although Cat-Conv consumes more parameters than DAFI, it causes performance decline when applied on Panini-Net, especially in detailed visual quality (\textit{e.g.}, tooth and eyes). We argue that's because the interpolation operation in DAFI can better preserve the details encapsulated in GAN Prior features, and the global condition guidance can help DAFI better handle feature fusion.     

\subsection{\textbf{B}: Dissection of Panini-Net}
To study the correlations between degradation levels and interpolation ratios, we fix $\sigma$, $\mathbf{\delta}$, $q$ in Eq.~\eqref{deg}, while choosing downsampling rate $r$ as 16, 32, 64, 128, respectively. Then we get four sets of degradation functions, the only difference between these degradation functions is the downsampling rate $r$. We use these four functions to process a single HQ image to get four degraded face images $\{\mathbf{X}_{\downarrow16}, \mathbf{X}_{\downarrow32}, \mathbf{X}_{\downarrow64}, \mathbf{X}_{\downarrow128}\}$ that share the same content but are different in degradation levels. We feed these four images into Panini-Net respectively. As Panini-Net uses progressive interpolation layer by layer, the earlier DAFI$_{i}$, $i\in\{1, ..., 5\}$, has weaker influences, while DAFI$_{5}$ has determinate affection on restoration. Hence, for each degraded face image, we only record the interpolation mask of DAFI$_{5}$. For simplicity, we sum the entire elements of the $\mathbf{mask}_{5}\in\mathbb{R}^{1\times 512}$ and divide it with the dimension to get a ratio $\mathbf{\theta}$ in range (0,1), defined as  
\begin{equation}
    \mathbf{\theta}=\sum_{j=1}^{512}\mathbf{mask}_{5}[j]/512.
\end{equation}
Obviously, $\mathbf{\theta}$ can coarsely reflect the usage ratio of features $\mathbf{F}_{GPB}^{5}$. The records are shown in Table \ref{ablation c maskval}. In DAFI$_{5}$, $\mathbf{\theta}$ shows obvious positive correlations with degradation levels: when degradation becomes severe, $\mathbf{\theta}$ will increase. That means Panini-Net tends to utilize more GAN Prior information when degradation becomes severe, which is in accord with our expectations.  
\begin{table}[t]
    \centering
        \caption{GAN Prior feature usage ratio in Panini-Net with different degradation levels. Panini-Net can dynamically increase the usage of GAN Prior when the degradation becomes severe.}
    \begin{tabular}{lcccc}
      \hline
         \rule{0pt}{10pt}{Downsampling Rate $r$} & 16 & 32 & 64 & 128\\
         \hline
         \rule{0pt}{10pt}{$\mathbf{F}_{GPB}^{5}$ usage ratio $\mathbf{\theta}$} &0.43&0.54&0.65&0.73\\
         \hline
    \end{tabular}
    \label{ablation c maskval}
\end{table}

Further, we expect to see what exactly does features $\mathbf{F}_{GPB}$ and $\mathbf{F}_{IFE}$ contribute to the final result when facing different degradation levels. We feed $\{\mathbf{X}_{\downarrow16}, \mathbf{X}_{\downarrow32}, \mathbf{X}_{\downarrow64}, \mathbf{X}_{\downarrow128}\}$ into Panini-Net respectively. At the first round, we get the standard results $\mathbf{Y}$, the operation of DAFI is the same as Eq.~\eqref{eq7}. At the second round, we manually set $\mathbf{F}_{GPB}$ as zero during inference to get results $\mathbf{Y}_{IFE}$, the operation of DAFI can be formulated as:
\begin{equation}
    \mathbf{F}_{DAFI}^{i} =\mathbf{F}_{IFE}^{i}\odot(\mathbf{1-mask}_{i}).
\end{equation}

At the third round, we manually set $\mathbf{F}_{IFE}$ as zero during inference to get results $\mathbf{Y}_{GPB}$, the operation of DAFI can be formulated as:
\begin{equation}
    \mathbf{F}_{DAFI}^{i} =\mathbf{F}_{GPB}^{i}\odot\mathbf{mask}_{i}.
\end{equation}

\begin{figure}[t]
  \centering
  \vspace{-2pt}
  \includegraphics[width=1\linewidth]{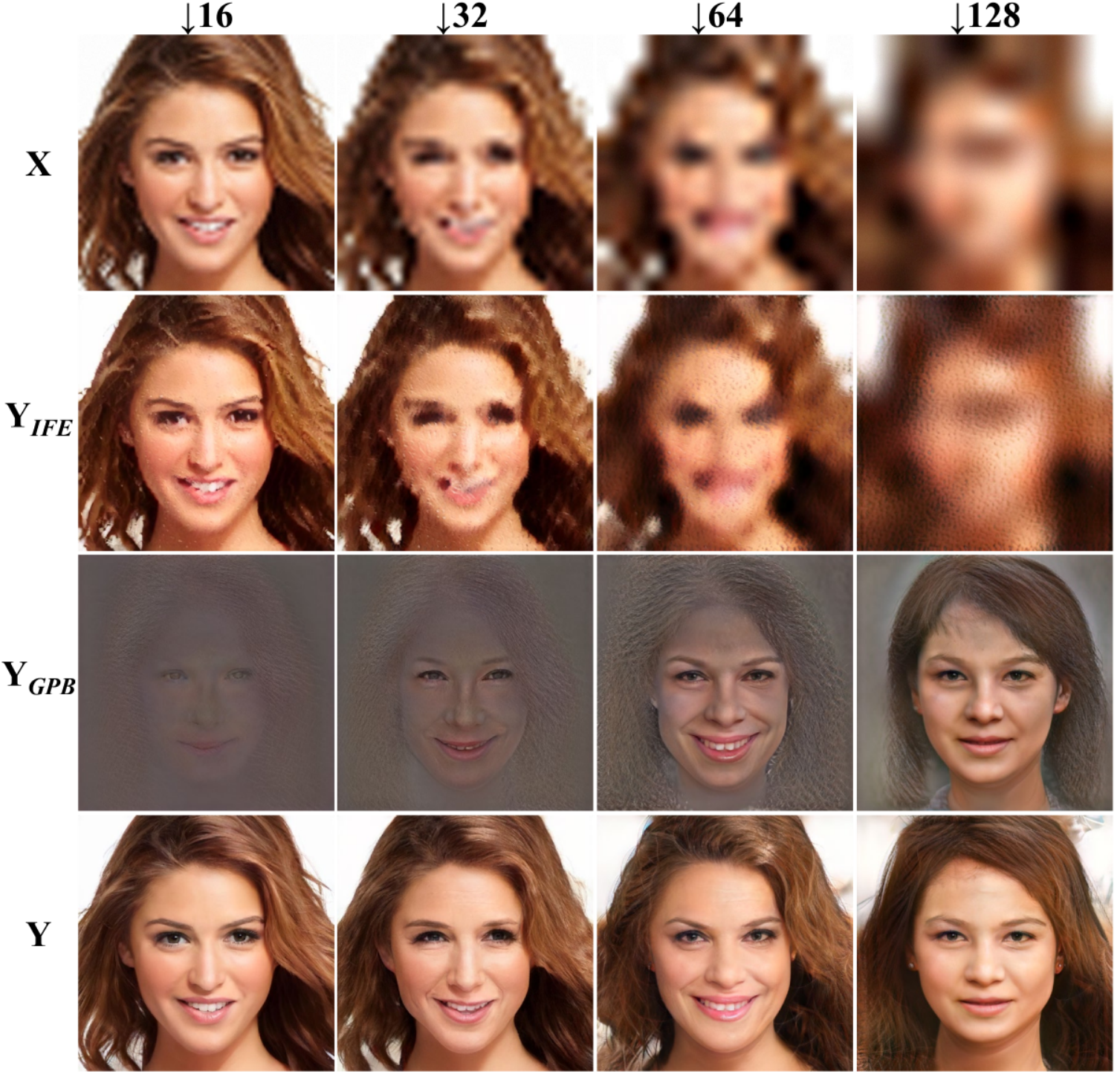}
  \caption{Ablation experiment \textbf{B}. Given four degraded face images $\mathbf{X}$ with downsampling rate 16, 32, 64, 128 respectively, $\mathbf{Y}$ are the standard results restored from $\mathbf{X}$ by Panini-Net. To solely use $\mathbf{F}_{GPB}$ for generation, we manually set $\mathbf{F}_{IFE}$ as $\mathbf{0}$ to generate the results $\mathbf{Y}_{GPB}$. To solely use $\mathbf{F}_{IFE}$ for generation, we manually set $\mathbf{F}_{GPB}$ as $\mathbf{0}$ to generate the results $\mathbf{Y}_{IFE}$. These results accord with our expectations for Panini-Net: (1) Dynamically fuse the two types of informative features (\textit{i.e.}, $\mathbf{F}_{IFE}$ and $\mathbf{F}_{GPB}$) with flexible adaption to various degradations. (2) Discriminatively utilize the rich details in GAN Prior and the overall structures of the degraded face image $\mathbf{X}$. \textbf{Zoom in for best view}.}
\label{ablation C picture} 
\vspace{-8pt}
\end{figure}

Results are shown in Fig.~\ref{ablation C picture}. When the degradation becomes severe, the valid contents of $\mathbf{Y}_{IFE}$ will decline, and $\mathbf{Y}_{GPB}$ will become more complete, which means Panini-Net can dynamically increase the use of GAN Prior to offset the decline of valid contents in the degraded face image. Besides, when the degradation is mild, the content of $\mathbf{Y}_{IFE}$ is mainly about the coarse structures of the input image, while the content of $\mathbf{Y}_{GPB}$ is mainly about the refinement of the eyes, nose, mouth, and textures. It indicates that Panini-Net learned how to extract valid information from GAN Prior and degraded face images. Another interesting discovery is that the masked features $\mathbf{F}_{GPB}^{i}\odot\mathbf{mask}_{i}$ and $\mathbf{F}_{IFE}^{i}\odot(\mathbf{1-mask}_{i})$ can generate reasonable results respectively, which implies the potential of DAFI for editing.  

\section{Conclusions}
In this paper, we present a novel feature fusion framework for face restoration, dubbed Panini-Net, which can dynamically fuse the features depending on degradation levels. The proposed DAFI provides a concise and efficient way to fuse external features into GAN prior. Besides, there is still a lot of room to explore the interpolation forms (\textit{e.g.} spatial-wise interpolation) and the way of mask generation. As we showed on the $16\times$ SR experiment, a simple modification on the way of mask generation can make Panini-Net competent to a new task, and we believe Panini-Net can also perform well in other supervised image-to-image tasks with limited modifications. 

On the other hand, the characteristics of interpolation operation enable us to explore the editability of the restored results. We can generate multiple high-quality solutions simply by adding a bias on mask elements (see appendix for more details). This merit brings more possibilities for practical applications and is worthy of study.

\appendix


\bibliography{aaai22}

\end{document}